\documentclass[journal]{IEEEtran}

\ifCLASSINFOpdf
\else
   \usepackage[dvips]{graphicx}
\fi
\usepackage{url}
\usepackage{graphicx}
\usepackage{amssymb, amsmath, arydshln, multirow, threeparttable}
\usepackage{amsfonts}
\usepackage{pifont}
\usepackage{booktabs}
\usepackage{diagbox}
\usepackage[colorlinks, linkcolor=blue, anchorcolor=blue, citecolor=blue]{hyperref}

\begin{document}

\title{Linguistic Steganalysis via LLMs: Two Modes for Efficient Detection of Strongly Concealed Stego}

\author{Yifan Tang, Yihao Wang, Ru Zhang*, Jianyi Liu
\thanks{This work is supported by the National Natural Science Foundation of China under Grant U21B2020.}
\thanks{Yifan Tang, Yihao Wang, Ru Zhang, and Jianyi Liu are with the School of Cyberspace Security, Beijing University of Posts and Telecommunications, Beijing 100876, China. (The corresponding author is Ru Zhang.) (E-mail: tyfcs@bupt.edu.cn, yh-wang@bupt.edu.cn, zhangru@bupt.edu.cn, and liujy@bupt.edu.cn)}
}
\markboth{}
{Shell \MakeLowercase{\textit{et al.}}: Bare Demo of IEEEtran.cls for IEEE Journals}
\maketitle

\begin{abstract}
To detect stego (steganographic text) in complex scenarios, linguistic steganalysis (LS) with various motivations has been proposed and achieved excellent performance. However, with the development of generative steganography, some stegos have strong concealment, especially after the emergence of LLMs-based steganography, the existing LS has low detection or cannot detect them. We designed a novel LS with two modes called LSGC. In the generation mode, we created an LS-task ``description" and used the generation ability of LLM to explain whether texts to be detected are stegos. On this basis, we rethought the principle of LS and LLMs, and proposed the classification mode. In this mode, LSGC deleted the LS-task ``description" and used the ``causalLM" LLMs to extract steganographic features. The LS features can be extracted by only one pass of the model, and a linear layer with initialization weights is added to obtain the classification probability. Experiments on strongly concealed stegos show that LSGC significantly improves detection and reaches SOTA performance. Additionally, LSGC in classification mode greatly reduces training time while maintaining high performance.
\end{abstract}


\IEEEpeerreviewmaketitle

\section{Introduction}

\IEEEPARstart{S}{teganography} is a technique for hiding systems in covert confidentiality systems \cite{intro}. It hides secret information in digital media such as texts \cite{rnn-stega} and images \cite{image}, generates steganographic media, and transmits them through public channels. Only authorized persons can detect whether the media is stego and accurately extract the secret. Thanks to the lossless transmission of text in social networks, linguistic steganography has been widely researched. The technology has evolved from early modified schemes easily detected \cite{tra-steganography}\cite{tra-steganography1} to generative schemes that automatically generate high-quality stegos \cite{gan-stega}\cite{vae-stega}. Recently, steganography based on open-source and closed-source large language models (LLMs) has been proposed \cite{llsm}, which shows extremely strong concealment and anti-detection performance. If these technologies are abused, security will be seriously endangered. Therefore, linguistic steganalysis (LS), as a risk prevention technology, has attracted the attention of scholars.

LS task aims to detect whether there is the secret in the texts. According to the design focus and model architecture, the existing LS is divided into traditional methods \cite{tra-steganalysis2} and deep-learning methods \cite{fcn}\cite{ts-rnn}. The former focuses on how to construct artificial features such as targeted word associations \cite{tra-steganalysis3} and use these features for detection. The latter focuses on how to design a deep-learning model that can extract high-dimensional features \cite{zou}. They use the model's representation to extract LS features with high diversity \cite{r-bilstm-c}. Since the stegos generated by generative steganography have high concealment, it is difficult for traditional methods to extract effective features, resulting in poor performance. Therefore, the research focus has shifted to the design of deep-learning LS.

\begin{table}[t]
	\vspace{-1ex}
	\centering
	\renewcommand\arraystretch{1.1}
	\setlength{\tabcolsep}{2.5mm}
	\caption{Summary of existing work, including whether to use pre-trained language models (PLM) and whether the stegos is LLMs-based (LLM stegos)}\label{intro}
	\vspace{-1ex}
	\begin{tabular}{ccccc}
		\toprule[1pt]
		Algorithms & PLM & Architecture & LLM stegos & Year \\
		\midrule[0.5pt]
		FCN \cite{fcn}				& \ding{55} & FCN				& \ding{55} & 2019 \\
		TS\_RNN \cite{ts-rnn}		& \ding{55} & RNN				& \ding{55} & 2019 \\
		R\_BI\_C \cite{r-bilstm-c}	& \ding{55} & LSTM, CNN 		& \ding{55} & 2019 \\
		TS\_GNN \cite{ts-gnn}		& \ding{55} & GNN				& \ding{55} & 2021 \\
		Zou \cite{zou}				& \ding{51} & LSTM				& \ding{55} & 2021 \\
		FS-stega \cite{few-shot} 	& \ding{51}	& LSTM  			& \ding{55} & 2022 \\
		MDA \cite{mda} 				& \ding{51} & CNN   			& \ding{55} & 2022 \\
		SSLS \cite{ssls}			& \ding{51} & CNN, GRU   		& \ding{55} & 2022 \\
		LS-LLL \cite{lll} 			& \ding{51} & K-means			& \ding{55} & 2022 \\
		Sesy \cite{sesy} 			& \ding{51} & GAT				& \ding{55} & 2022 \\
		LS\_HML \cite{ins} 			& \ding{55} & Mutual learning 	& \ding{55} & 2022 \\
		RLS\_DTS \cite{rls-dts} 	& \ding{55} & Actor-Critic		& \ding{55} & 2023 \\
		LSFLS \cite{lsfls} 			& \ding{51} & BNN   			& \ding{55} & 2023 \\
		UP4LS \cite{up4ls} 			& \ding{51} & User profile 		& \ding{55} & 2023 \\
		SANet \cite{SANet}			& \ding{51} & CNN 				& \ding{55} & 2024 \\
		GS-Llama \cite{gs-Llama} 	& \ding{51} & LLMs   			& \ding{55} & 2024 \\
		\midrule[0.5pt]
		\textbf{Ours (LSGC)} 		& \ding{51} & LLMs  			& \ding{51} & 2024 \\
		\bottomrule[1pt]
	\end{tabular}%
	\vspace{-4ex}
\end{table}%

\begin{figure*}[!htbp]
	\vspace{-1ex}
	\centering
	\includegraphics[width=0.97\textwidth]{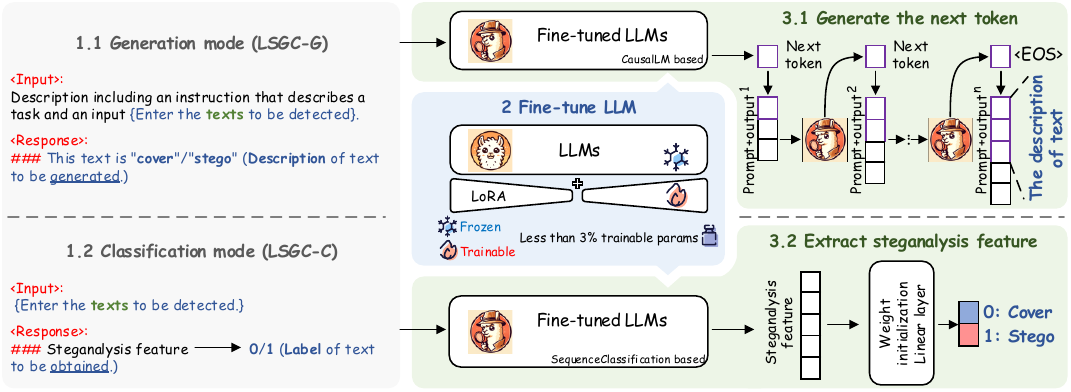}
	\caption{The overall framework of the LSGC method. The entire framework mainly consists of two modes: generation and classification. The ``Prompt" of the generation mode requires the construction of the ``\textbf{Description}" and ``\textbf{Instruction}" (LS-task), the ``\textbf{Input}" (texts to be detected), and the fine-tuned LLMs is used to obtain the ``Response" of the next token. This cycle repeats until the next token is the stop symbol ``$<$EOS$>$", and the generated content is the description of whether the text is stego. The input of the classification mode deletes the LS-task ``Description" and only requires ``Instruction". The fine-tuned LLMs are used to obtain the steganalysis features, and then a linear layer with initial random weights is used to map the features to label probabilities.}\label{fig1}
	\vspace{-3ex}
\end{figure*}

In recent years, a series of representative LS works have emerged. To use the dependency between words with a long distance for LS, Yang et al. \cite{ts-rnn} and Zou et al. \cite{zou} designed LS models based on RNN and LSTM architectures. These models extracted the correlation between words and achieved good performance in ideal datasets. Niu et al. \cite{r-bilstm-c} combined the advantages of LSTM and CNN, Xu et al. \cite{ssls} combined the advantages of CNN and GRU to explore a hybrid LS and achieved better performance. To extract the difference in conditional probability distribution caused by LS embedding, Wu et al. \cite{ts-gnn} proposed a method based on GNN, which improved the detection. Wang et al. \cite{lsfls} and Wen et al. \cite{few-shot} invented a self-training and meta-learning-based LS method respectively, and performed effective detection in few-sample scenarios. Faced with different domains, ordinary LS finds it difficult to detect stegos in cross-domain data. Wen et al. \cite{lll}, Xue et al. \cite{mda}, and Wang et al. \cite{rls-dts} successively proposed cross-domain LS based on lifelong learning, domain adaptation, and reinforcement learning, and achieved excellent detection. Bai et al. \cite{gs-Llama}, who proposed this work at a similar time, showed the performance of LLMs in LS tasks and achieved better performance. In addition, there are some works dedicated to the design of frameworks. Xue et al. \cite{ins} constructed a framework based on hierarchical mutual learning and \cite{SANet} proposed adaptive domain-invariant feature extraction method, Yang et al. \cite{sesy} captured contextual features, and Wang et al. \cite{up4ls} constructed user profiles to extract user features. These works improve the performance or efficiency of existing LS in complex scenarios. TABLE \ref{intro} summarizes the design details of existing works.

Although the above methods have good results for most generative stegos, they still have the following problems: most LS methods are small in scale and exhibit poor performance on stegos with strong concealment. Large-scale models show excellent performance but also bring huge training costs. Therefore, the need for efficient LS of strongly concealed stegos needs to be addressed urgently. In this letter, we re-examine the principle of LLMs and the essence of LS-task, and propose the LSGC method with two modes: generation and classification. The LLMs used are Llama2 and Llama3. Extensive experiments show that LSGC significantly improves performance in hard-to-detect datasets and reaches SOTA performance. LSGC with classification mode also reduces the training time compared to the SOTA baseline. Additionally, we explore the relationship between the scale of the fine-tuned model and detection performance.

The main contributions of this letter are as follows.

\begin{enumerate}
	\item In the generation mode of LSGC, the Prompt paradigm of LLMs is used. We describe the LS-task and use it as input together with the training text, and the output is the conditional probability distribution of the next token. This mode of LSGC can get a description of whether the text is stego.
	
	\item In the classification mode of LSGC, the working principle of LLMs is rethought. We delete the LS-task description in the conventional LLMs and reduce the length of the input sequence. Then, we add an initialized weights linear layer to LLMs to convert the extracted steganalysis features into probabilities. The input is modified to a label, thus avoiding the loop process of the conventional LLM output terminator.
	
	\item Experiments on strongly concealed data show that the proposed LSGC surpasses the baselines, and also takes less training time than the SOTA baseline.
\end{enumerate}

The rest of this letter is arranged as follows: Section \ref{sec2} introduces the model details of the LSGC method proposed in detail. Through a large number of experiments in the strongly concealed data, Section \ref{sec3} gives a comparison and result analysis of LSGC and the baselines. Finally, Section \ref{sec4} summarizes the letter.

\vspace{-1ex}
\section{Proposed LSGC}\label{sec2}
In this section, we will describe the proposed LSGC with two modes, generation mode (LSGC-G) and classification mode (LSGC-C) in detail. Its overall framework is shown in Fig. \ref{fig1}.

\vspace{-2ex}
\subsection{Fine-tuning of LLMs}\label{subsec21}
Since the cost of fine-tuning all parameters of a large model is too high, we adopt the LoRA strategy \cite{lora} to freeze the internal parameters of LLM and construct a low-rank (i.e., low-dimensional) matrix outside LLM for training. The number of parameters in this low-rank matrix is much smaller than that of the LLMs itself. The formula is shown below. 
\begin{equation}\label{eq1}
{{\bf{W}}_0} + \Delta {\bf{W}} = {{\bf{W}}_0} + BA,B \in {\mathbb{R}^{d \times r}},A \in {\mathbb{R}^{r \times k}},
\end{equation}
\noindent where, $r\ll\min(d,k)$. By merging the trained low-rank matrix with the original LLM parameters, efficient fine-tuning of large models can be achieved while retaining the original pre-training knowledge.

\vspace{-2ex}
\subsection{Generation Mode (LSGC-G)}\label{subsec22}
By training the LLMs' understanding ability, the LLMs can generate outputs corresponding to understood text content based on understanding the input text. This letter uses the LLMs' understanding ability to generate an explanation of whether the input text is stego, completing the LS task.

The input of the LSGC-G needs to construct a Prompt. It includes the relevant ``Description" and ``Instruction" for the LS task, an ``Input" for the text to be detected, and a blank ``Response". This ``Description" allows the LLMs to understand the direction that needs to be generated. The conditional probability distribution of the next token is generated using the fine-tuned ``CausalLM" LLM to determine the next token. Then the next token will be added to the Prompt and input into the LLM again. Repeatedly until the stop symbol ``$<$EOS$>$" is generated. The formula is as follows.
\begin{equation}\label{eq2}
\small
\begin{array}{l}
{\rm output}{^i} = LLM({\rm Prompt} + {\rm output}^1 + \cdots + {\rm output}^{i - 1})\\
if\ {\rm output}^{i - 1} \ne <{\rm{EOS}}> ,
\end{array}
\end{equation}
\noindent where all the generated tokens are connected to describe whether the input text is stego.

\vspace{-2ex}
\subsection{Classification Mode (LSGC-C)}\label{subsec23}
Based on the required input of LLMs and the working principle of LS models, we re-examine the shortcomings of the generation mode and GS-Llama \cite{gs-Llama} in LS tasks.

First, in terms of model input, the input of the conventional LS model is the text to be detected. However, the generation mode and GS-Llama require an additional task-related ``Description". This increases the length of the sequence received by the model, and the sequence length of the ``Description" is even much longer than the sequence length of the text to be detected. Second, in terms of model output results, the conventional LS model only needs to pass once to obtain the extracted features. However, the generation mode and GS-Llama need to generate a description of the text to be detected, which requires multiple inputs into LLMs to obtain the next token. Even if only one word is generated, such as ``cover" or ``stego", it is necessary to enter LLMs again to get the stop symbol ``$<$EOS$>$". They greatly increase the training time.

Therefore, we constructed this classification mode. We deleted the ``Description" in the LSGC-G and converted the LLM of the ``CausalLM" to the ``SequenceClassification" architecture. The LS features are obtained by the ``SequenceClassification" LLMs. The formula is shown below.
\begin{equation}\label{eq3}
{{\bf{E}}^L} = Tr{m_{enc}}({{\bf{E}}^{L - 1}}),
\end{equation}
\noindent where, ${{\bf{E}}^{L - 1}}$ is both the $L-1$th layer output vector and the $L$th layer input vector. Then a linear layer with initial random weights is added to get the probability of the final label. This can significantly reduce the training time while ensuring the detection performance.

\vspace{-1ex}
\section{Experiments}\label{sec3}
This section shows the performance comparison of the LSGC method and the baselines. To ensure the fairness and reliability of the comparison, each data is repeated 5 times and averaged. All experiments are run on the NVIDIA GeForce RTX 3090 GPUs.

\vspace{-2ex}
\subsection{Experiment Settings}\label{subsec41}

\paragraph* {\textbf{Dataset}} In terms of dataset selection, we used VAE-Stega \cite{vae-stega} and LLsM \cite{llsm} steganographic schemes. VAE-Stega uses three classic text sets Movie, News, and Twitter for model training, and LLsM uses high-quality text generated by GPT4 as fine-tuning data to generate corresponding stegos. To simulate real steganographic samples, we used regular expressions to filter out situations where the text was too long, too short, and garbled, increasing the difficulty of the detection task. The entire dataset is divided into training, validation, and testing sets in a ratio of 6:2:2. And we set the same amount of 1000 for the cover texts and stego texts. TABLE \ref{dataset} describes the dataset used.

\begin{table}[t]
	\vspace{-1ex}
	\centering
	\renewcommand\arraystretch{1.1}
	\caption{Detailed description of used two dataset.}\label{dataset}
	\vspace{-1ex}
	\begin{tabular}{ccccc}
		\toprule[1.2pt]
		\multicolumn{1}{c}{\textbf{Schemes}} & \multicolumn{2}{c}{\textbf{Setting \& Data}} & \multicolumn{1}{c}{\textbf{BPW}} & \multicolumn{1}{c}{\textbf{Total Texts}} \\
		\midrule[0.5pt]
		\multicolumn{1}{c}{\multirow{3}{*}{VAE-Stega \cite{vae-stega}}} & \multicolumn{1}{c}{\multirow{3}{*}{AC / HC}} & Movie & \multirow{3}{*}{5} & \multicolumn{1}{c}{\multirow{3}{*}{12,000}} \\
		&       & Twitter	&       &  \\
		&       & News  	&       &  \\
		\midrule[0.5pt]    
		
		\multicolumn{1}{c}{\multirow{3}{*}{LLsM \cite{llsm}}} & \multicolumn{1}{c}{\multirow{3}{*}{$\alpha= 8/16/ 32$}} & \multirow{1.7}{*}{$\beta=1$}   & \multirow{1.7}{*}{1}   &  \multicolumn{1}{c}{\multirow{3}{*}{12,000}}\\
		&       & \multirow{2.5}{*}{$\beta=0.5$} & \multirow{2.5}{*}{2}  &  \\
		&       &   		&       &  \\

		\bottomrule[1.2pt]
	\end{tabular}%
	\vspace{-3ex}
\end{table}%

\vspace{0.5ex}
\paragraph* {\textbf{Baselines}} We selected 6 high-performance methods as baselines and compared them with the LSGC method. These baselines include: \textbf{Non-BERT-based}: FCN \cite{fcn}, R\_BI\_C \cite{r-bilstm-c}. \textbf{BERT-based}: Zou \cite{zou}, Sesy \cite{sesy}, SSLS \cite{ssls}. \textbf{LLMs-based}: GS-Llama \cite{gs-Llama}.
\vspace{0.5ex}
\paragraph* {\textbf{Hyperparameters}} In the LSGC method, batch\_size is set to 10, the learning rate is initialized to 5e-5, the learning algorithm is the AdamW algorithm \cite{adamw}, the epoch is set to 5, $r$ in LoRA \cite{lora} is set to 64, $lora\_\alpha$ is set to 128 which usually set twice as much as $r$, and $lora\_dropout$ is set to 0.5. In the comparison experiment, LSGC used Llama2-7B \cite{Llama2} LLM. In the ablation experiment, two lastest models Llama2-7B and Llama3-8B \cite{Llama3} were used to discuss the effects of different settings and base-model.
\vspace{0.5ex}
\paragraph* {\textbf{Evaluation Metrics}} We use the detection accuracy Acc and F1 score to evaluate the detection performance of the methods. They were calculated by: $Acc = \frac{{{TP + TN}}}{{{TP + FP + TN + FN}}}$ and $F1 = \frac{2 \times (P \times R)}{(P + R)}$, Where, TP, TN, FP, and FN represent true positive, true negative, false positive, and false negative. $P$ and $R$ represent the Precision and Recall of detection. We use the time consumption $Time$ to evaluate the training and reasoning speed of the model,

\begin{table*}[!htbp]
	\vspace{-2ex}
	\centering
	\renewcommand\arraystretch{1.1}
	\setlength{\tabcolsep}{2.1mm}
	\caption{Reported accuracy(\%) and F1-score(\%) of two datasets. \textbf{Bold values} represent the best results, and \underline{underlined values} represent the suboptimal results.}\label{tab3}%
	\vspace{-1ex}
	\begin{tabular}{cccccccccccccc}
		\toprule[1.2pt]
		\multicolumn{2}{c}{\multirow{4}*{\diagbox[width=11em]{\textbf{Steganalysis}}{\textbf{Steganography}}}} & \multicolumn{6}{c}{VAE-Stega \cite{vae-stega}} & \multicolumn{6}{c}{LLsM \cite{llsm}} \\
		\multicolumn{2}{c}{} & \multicolumn{3}{c}{\multirow{2}*{AC}} & \multicolumn{3}{c}{\multirow{2}*{HC}} & \multicolumn{2}{c}{$\alpha=8$} & \multicolumn{2}{c}{$\alpha=16$} & \multicolumn{2}{c}{$\alpha=32$}\\
		\multicolumn{8}{c}{}{} & $\beta=1$ & $\beta=0.5$ & $\beta=1$ & $\beta=0.5$ & $\beta=1$ & $\beta=0.5$ \\
		\cdashline{3-14}[3pt/3pt]
		\multicolumn{2}{c}{}{} & Movie & Twitter & News & Movie & Twitter & News & LLsM & LLsM & LLsM & LLsM & LLsM & LLsM \\
		\midrule[0.5pt]
		\multirow{2}{*}{FCN \cite{fcn}} & $Acc$   & 57.63  & 53.75  & 52.55  & 61.63  & 57.75  & 74.38  & 54.50  & 51.37  & 51.60  & 56.97  & 52.35  & 59.50  \\
		& $F1$    & 50.65  & 34.63  & 46.31  & 60.89  & 66.65  & 73.89  & 47.09  & 47.55  & 49.47  & 52.35  & 49.20  & 57.70 \\
		\cdashline{1-14}[3pt/3pt]
		
		\multirow{2}{*}{R-BiLSTM-C \cite{r-bilstm-c}} & $Acc$   & 64.50  & 58.75  & 62.50  & 87.63  & 80.75  & 94.38  & 64.75  & 61.05  & 61.88  & 67.42  & 63.83  & 68.73  \\
		& $F1$    & 66.91  & 56.46  & 64.95  & 88.03  & 82.14  & 94.33  & 63.85  & 61.05  & 67.03  & 66.16  & 63.82  & 68.83  \\
		\cdashline{1-14}[3pt/3pt]
		
		\multirow{2}{*}{Zou \cite{zou}} & $Acc$   & 87.38  & 74.03  & 92.25  & 92.75  & 86.81  & 97.38  & 70.63  & 65.43  & 65.06  & 65.87  & 65.56  & 73.25  \\
		& $F1$    & 87.39  & 74.21  & 91.99  & 93.21  & 86.85  & 97.35  & 70.44  & 61.48  & 65.28  & 65.15  & 67.53  & 74.27  \\
		\cdashline{1-14}[3pt/3pt]
		
		\multirow{2}{*}{SSLS \cite{ssls}} & $Acc$   & 90.75  & 78.75  & 95.25  & 95.25  & 88.38  & 98.13  & 70.38  & 65.50  & 64.38  & 71.12  & 65.50  & 74.25  \\
		& $F1$    & 90.74  & 78.75  & 95.25  & 95.25  & 88.36  & 98.12  & 70.34  & 65.50  & 64.27  & 70.99  & 65.31  & 74.25 \\
		\cdashline{1-14}[3pt/3pt]
		
		\multirow{2}{*}{Sesy \cite{sesy}} & $Acc$   & 92.50  & 75.38  & 95.88  & 94.75  & 88.32  & 97.62  & 70.12  & 64.25  & 64.25  & 73.75  & 66.50  & 73.88  \\
		& $F1$    & 92.19  & 73.20  & 95.62  & 94.53  & 88.26  & 97.60  & 72.31  & 66.19  & 66.29  & 71.47  & \underline{70.02}  & 69.04 \\
		\cdashline{1-14}[3pt/3pt]
		
		\multirow{2}{*}{GS-Llama \cite{gs-Llama}} & $Acc$   & 92.63  & 78.13  & 95.13  & 94.63  & 89.38  & 96.75  & 68.38  & 63.63  & 65.44  & 74.25  & 65.38  & 72.13  \\
		& $F1$    & 92.59  & 78.06  & 95.12  & 94.62  & 89.13  & 96.75  & 68.24  & 63.15  & 65.59  & 74.13  & 65.15  & 72.86 \\
		\midrule[0.8pt]
		
		\multirow{2}{*}{\textbf{Ours(LSGC-G)}} & $Acc$ & \textbf{97.44} & \textbf{92.75} & \underline{97.94} & \underline{97.45} & \underline{93.44} & \textbf{99.13} & \underline{74.73} & \underline{72.13} & \underline{70.63} & \textbf{82.25} & \textbf{70.88} & \textbf{83.63} \\
		& $F1$    & \textbf{98.04}  & \underline{92.95}  & \underline{97.54}  & \textbf{97.84}  & \textbf{93.74}  & \textbf{99.10}  & \underline{74.92}  & \underline{72.39}  & \underline{70.57} & \textbf{82.14}  & \textbf{70.74}  & \textbf{83.32}  \\
		\cdashline{1-14}[3pt/3pt]
		
		\multirow{2}{*}{\textbf{Ours(LSGC-C)}} & $Acc$   & \underline{97.38}  & \underline{92.38}  & \textbf{98.38}  & \textbf{98.13}  & \textbf{93.88}  & \underline{98.88}  & \textbf{75.38} & \textbf{72.63}  & \textbf{71.75}  & \underline{81.75}  & \underline{69.38}  & \underline{83.25}  \\
		& $F1$    & \underline{96.87}  & \textbf{92.97}  & \textbf{98.57}  & \underline{97.10}  & \underline{92.97}  & \underline{98.97}  & \textbf{75.27}  & \textbf{72.41}  & \textbf{71.35}  & \underline{81.55}  & 69.32  & \underline{83.15}  \\
		\bottomrule[1.2pt]
	\end{tabular}%
	\vspace{-3ex}
\end{table*}%

\vspace{-2ex}
\subsection{Experiment Performance Comparison}\label{subsec42}

The detection comparison of the proposed LSGC method and the baselines in different datasets is shown in TABLE \ref{tab3}.

According to TABLE \ref{tab3}, the LSGC performance significantly surpasses the BERT-based and LLM-based baselines. It is not difficult to find that the LLM-based baseline \cite{gs-Llama} does not surpass some BERT-based baselines. This is because the scale of its fine-tuned model is too small, which limits the potential of LLMs in LS. For the impact of the scale of the fine-tuned model on performance, see the Section \ref{subsec44}.

\vspace{-2ex}
\subsection{Experiment Performance in Mixed Scenarios}\label{subsec43}

We design three mixed scenarios: a mixed stegos by different corpora using VAE-Stega, a mixed stegos by different embedding rates using LLsM, and a mixed stegos by VAE-Stega and LLsM. The results are shown in TABLE \ref{tab4}. Our proposed two modes of LSGC achieve the best result.

\begin{table}[t]          
	\centering
	\renewcommand\arraystretch{1.1}
	\setlength{\tabcolsep}{3.4mm}
	\caption{Reported accuracy(\%) and F1-score(\%) in mixed scenarios.}\label{tab4}%
	\vspace{-1ex}
	\begin{tabular}{ccccc}
		
		\toprule[1.2pt]	\multicolumn{2}{c}{\diagbox[width=11em]{\textbf{Steganalysis}}{\textbf{Steganography}}} & {\cite{vae-stega}-mix} & {\cite{llsm}-mix} & {\cite{vae-stega}+\cite{llsm}} \\
		\midrule[0.5pt]
		\multirow{2}{*}{FCN \cite{fcn}} & $Acc$	& {63.08}	& {57.55} & {69.13} \\
		{} & $F1$ & {63.84}  & {57.25}  & {71.95}  \\
		
		\cdashline{1-5}[3pt/3pt]
		
		\multirow{2}{*}{R-BiLSTM-C \cite{r-bilstm-c}} & $Acc$ & {76.88}	& {62.40}  & {72.63} \\
		{} & $F1$ & {73.95} & {62.92} & {73.71} \\
		
		\cdashline{1-5}[3pt/3pt]
		
		\multirow{2}{*}{Zou \cite{zou}} & $Acc$ & {82.63} & {59.50} & {77.13} \\
		{} & $F1$ & {83.89} & {60.80} & {78.08}\\
		
		\cdashline{1-5}[3pt/3pt]
		
		\multirow{2}{*}{SSLS \cite{ssls}} & $Acc$ & {89.63} & {59.63} & {75.63} \\
		{} & $F1$ & {89.61} & {59.10} & {75.36} \\
		
		\cdashline{1-5}[3pt/3pt]
		
		\multirow{2}{*}{Sesy \cite{sesy}} & $Acc$ & {87.62} & {60.50} & {81.25} \\
		{} & $F1$ & {87.39} & {59.07} & {80.67} \\
		
		\cdashline{1-5}[3pt/3pt]
		
		\multirow{2}{*}{GS-Llama \cite{gs-Llama}} & $Acc$ & {91.88} & {63.38} & {77.13} \\
		{} & $F1$ & {91.74} & {63.24} & {77.10} \\
		
		\midrule[0.8pt]
		
		\multirow{2}{*}{\textbf{Ours(LSGC-G)}} & $Acc$ & {\underline{95.63}} & {\underline{66.70}} & {\textbf{82.50}} \\
		{} & $F1$	& {\underline{95.62}} & {\underline{66.74}} & {\textbf{82.51}} \\
		
		\cdashline{1-5}[3pt/3pt]
		
		\multirow{2}{*}{\textbf{Ours(LSGC-C)}} & $Acc$ & {\textbf{96.75}} & {\textbf{67.88}} & {\underline{82.38}} \\
		{} & $F1$ & {\textbf{96.75}} & {\textbf{67.84}} & {\underline{82.20}} \\
		
		\bottomrule[1pt]
	\end{tabular}
	\vspace{-1ex}
\end{table}

\begin{table}[t]
	\vspace{-1ex}
	\centering
	\renewcommand\arraystretch{1.1}
	\setlength{\tabcolsep}{1.5mm}
	\caption{Training time of LLMs-based methods.The time unit is minute.}\label{tab5}%
	\begin{tabular}{cccc}
		\toprule[1.2pt]
		\multicolumn{1}{c}{\textbf{Steganalysis}} & \multicolumn{1}{c}{GS-Llama \cite{gs-Llama}} & \multicolumn{1}{c}{LSGC-G} & \multicolumn{1}{c}{LSGC-C} \\
		\midrule[0.5pt]
		\textbf{$Time$} & 33.72 & 28.95 ($\downarrow$ 14.15\%) & 14.34 ($\downarrow$ 57.47\%) \\
		\bottomrule[1.2pt]
	\end{tabular}%
	\vspace{-2ex}
\end{table}%

Since the effects of Non-BERT-based and BERT-based are weak, we compare the training time with the LLM-based baseline \cite{gs-Llama}. The comparison results are shown in TABLE \ref{tab5}. We find that LSGC-C can reduce the training time. LSGC-G takes a shorter training time, while LSGC-C only requires half time as it does not require an excessively long prompt.

\vspace{-2ex}
\subsection{Ablation Experiments}\label{subsec44}
We also explore the effect of different LLMs and the scale of fine-tuned model on performance in LLsM-mix dataset. The results are shown in TABLE \ref{tab6}. According to Table \ref{tab6}, it can be found that the advantages of different LLMs in LS tasks are not obvious. On the contrary, $r$, which determines the scale of the fine-tuning model, plays a decisive role in the LS performance. This shows that the more training parameters there are in complex steganalysis tasks, the better the effect. We also attempted to investigate the impact of different prompts on the generative expression, but the results showed that the impact was minimal.

\begin{table}[t]
	\centering
	\renewcommand\arraystretch{1.1}
	\setlength{\tabcolsep}{3mm}
	\caption{Ablation experiments with different models and scales.}\label{tab6}%
	\vspace{-1ex}
	\begin{tabular}{cccccc}
		\toprule[1.2pt]
		\multirow{2}{*}{\textbf{Base-Models}} & \multirow{2}{*}{Setting} & \multicolumn{2}{c}{LSGC-G} & \multicolumn{2}{c}{LSGC-C} \\
		\cline{3-6}          &       & $Acc$  & $F1$    & $Acc$   & $F1$ \\
		\midrule[0.5pt]
		\multirow{5}{*}{Llama2-7B} 
		& $r=8$   &58.13 &	57.51 &	65.44 &	65.44   \\
		& $r=16$  & 62.38 &	62.37 &	66.31& 	66.29   \\
		& $r=32$  & 65.50  & 65.48  & 67.38  & 67.36  \\
		& $r=64$  & 66.70  & 66.74  & 67.88  & 67.84  \\
		& $r=128$ & 66.50  & 66.50  & 66.06  & 66.06  \\
		
		\bottomrule[0.5pt]
		
		\multirow{5}{*}{Llama3-8B} & $r=8$ & 60.18  & 59.91  & 66.14  & 66.08  \\
		& $r=16$ & 63.88  & 62.75  & 66.77  & 66.47  \\
		& $r=32$ & 65.93  & 66.04  & 67.18  & 66.91  \\
		& $r=64$ & 67.88  & 67.84  & 68.18  & 67.74  \\
		& $r=128$ & 67.33  & 67.19  & 66.98  & 66.56  \\
		\bottomrule[1.2pt]
	\end{tabular}%
	\vspace{-3ex}
\end{table}%

\vspace{-1ex}
\section{Conclusion}\label{sec4}
To enhance the detection in strongly concealed stegos, this letter proposes the LSGC method with two modes. This method designs the generation and classification mode by examining the working principle of LS and the essence of generative LLM generation. Experiments show that the performance of LSGC surpassed the BERT-based and LLM-based baselines. At the same time, LSGC-C greatly reduces the training time than LLM-based baselines.


\end{document}